\documentclass{article}

\usepackage{geometry}
\usepackage[numbers]{natbib}

\usepackage{amsmath} 
\usepackage{amssymb}
\usepackage{latexsym}
\usepackage[ruled,linesnumbered]{algorithm2e}
\usepackage{dsfont}

\usepackage{float}
\usepackage{caption}
\usepackage{inputenc}
\usepackage{epsfig}
\usepackage{multicol}
\usepackage{graphicx}
\usepackage{url}
\usepackage{subfigure}
\usepackage{multirow}

\DeclareMathOperator*{\argmax}{arg\,max}

\begin{document}

\title{An Optimal Bayesian Network Based Solution Scheme for the Constrained Stochastic On-line Equi-Partitioning Problem\footnote{A preliminary version of this paper was presented at ICMLA 2014 - the 13th International Conference on Machine Learning and Applications, Detroit, USA, December 2014.
}}

\author{Sondre Glimsdal\thanks{This author can be contacted at: Centre for Artificial Intelligence Research (CAIR), University of Agder, Postbox 422, 4604 Kristiansand, Norway.  E-mail: {\tt
sondre.glimsdal@uia.no}.} ~and~Ole-Christoffer Granmo\thanks{Author's status: {\it Professor}. This author can be contacted at: Centre for Artificial Intelligence Research (CAIR), University of Agder, Postbox 422, 4604 Kristiansand, Norway.  E-mail: {\tt
ole.granmo@uia.no}.}}

\date{}

\maketitle

\begin{abstract}
A number of intriguing decision scenarios revolve around partitioning a
collection of objects to optimize some application specific objective
function. This problem is generally referred to as the Object Partitioning
Problem (OPP) and is known to be NP-hard. We here consider a particularly
challenging version of OPP, namely, the Stochastic On-line Equi-Partitioning
Problem (SO-EPP). In SO-EPP, the target partitioning is unknown and has to be
inferred purely from observing an on-line sequence of object pairs. The paired
objects belong to the same partition with probability $p$ and to different partitions with probability $1-p$, with $p$
also being unknown. As an additional complication, the partitions are required to be
of equal cardinality. Previously, only sub-optimal solution strategies have been proposed for SO-
EPP. In this paper, we propose the first optimal solution strategy. In brief,
the scheme that we propose, BN-EPP, is founded on a Bayesian network
representation of SO-EPP problems. Based on probabilistic reasoning, we are
not only able to infer the underlying object partitioning with optimal
accuracy. We are also able to simultaneously infer $p$, allowing us to
accelerate learning as object pairs arrive. Furthermore, our scheme is the
first to support arbitrary constraints on the partitioning (Constrained SO-EPP).
Being optimal, BN-EPP provides superior performance compared to existing
solution schemes.  We additionally introduce Walk-BN-EPP, a novel WalkSAT
inspired algorithm for solving large scale BN-EPP problems. Finally, we
provide a BN-EPP based solution to the problem of order picking,
a representative real-life application of BN-EPP.
%
\end{abstract}

%
%
\section{Introduction}
\label{sec:introduction}
A number of intriguing decision scenarios revolve around grouping a collection
of objects into partitions in such a manner that some application specific
objective function is optimized. This type of grouping is referred to as the
Object Partitioning Problem (OPP) and is in its general form known to be NP-hard.

In this paper, we consider a particularly challenging variant of OPPs
--- the Constrained Stochastic Online Equi-Partitioning Problem (CSO-EPP). In
    CSO-EPP, objects arrive sequentially, in pairs.  Furthermore, the
    relationship between the arriving objects is stochastic: Paired objects
    belong to the same partition with probability $p$, and to different ones
    with probability $1-p$. As an additional complication, the partitioning is
    constrained, with the default constraint being that the partitions must be
    of equal cardinality, referred to as equi-partitioning. Under these
    challenging conditions, the overarching goal is to infer the underlying
    partitioning, that is, to predict which objects will appear together in
    future arrivals, from a history of object arrivals.

The CSO-EPP can be applied to solve a number of challenging
tasks. We will here study a particularly fascinating one,
\emph{order picking}, which highlights the full spectrum of nuisances captured by
CSO-EPP. Order picking is defined as \emph{"the process of retrieving products
from storage (or buffer areas) in response to a specific customer
request"} \cite{deKoster2007}. Order picking occurs both in warehouses
employing an Automated Storage/Retrieval System (AS/RS) and those depending on
manual labor. Tompkins et al. identified travel time as the main factor when
it comes to optimizing order-picking \cite{tompkins2010}. For this reason, to
facilitate efficient retrieval of products, frequently ordered products should
be placed in easy to reach locations. Additionally, products that are often
ordered together should be placed in near-proximity of each other. By
doing so, we can systematically reduce the total travel time needed to collect orders.

In more challenging order-picking scenarios, the governing product
relationships may be unknown initially, and thus have to be learned over time
by monitoring which products are ordered together. Additionally, non-related
products may sporadically be ordered in conjunction, leading to
\emph{stochastic} order composition. This means that successful solution
strategies must be able to operate in a stochastic environment. Furthermore, many order picking scenarios impose constraints
when it comes to product placement. One could for instance require that a
subset of the objects is located in a subset of the available locations, e.g.,
that all frozen objects should be in freezers, even when they are rarely
purchased together. Other constraints could be that all products from a brand
must be co-located on the request of the manufacturer, or
that fragile objects must be placed in shelves close to the floor. To further exemplify the
importance of dealing with constraints, several more are listed in
Table \ref{tab:warehouse-restrictions}\footnote{These are based on real-world
point-of-sale transaction data from a grocery outlet \cite{hahsler2006implications}.}. Noting that each section of a warehouse can be represented as a CSO-EPP
partition, and that products can be represented as CSO-EPP
objects, we propose CSO-EPP as a model for order picking.

\begin{table}[]
\centering
\caption{Example constraints governing the placement of products in a warehouse.}
\label{tab:warehouse-restrictions}
\begin{tabular}{|l|l|l|}
\hline
Number & Products                & Constraint                                  \\ \hline
1 & shopping bags & Must either be in the entrance- or counter section \\ \hline
2 & whole milk, rolls/buns, tropical fruit  & Cannot be in the same section       \\ \hline
3 & white wine, specialty chocolate & Must be in the same section \\ \hline
4 & yogurt           & Has to be in the cooler section \\ \hline
5 & tropical fruit & Cannot be in the cooler section \\ \hline
\end{tabular}
\end{table}

In this paper, we present the first {\it optimal} solution scheme for SO-EPP and CSO-EPP. Let $\mathcal{O} = \{O_1, O_2,$ $\ldots, O_w \}$ be
a set of $W$ objects. These are to be partitioned into $R$ different
partitions $\mathcal{P} = \{P_1, P_2, \ldots, P_R\}$. The aim is to find
some unknown underlying partitioning of the objects based on noisy observations. Succinctly, the problem can be described as a 2-tuple $(\mathcal{U}, p)$,
where $\mathcal{U}$ is a set of tuples $(O_i, O_j)$. If $(O_k, O_m) \in \mathcal{U}$ then object $k$ and $m$ belong to the
same underlying partition, otherwise, they belong to different ones. Constraints can then naturally be formulated in terms of: (1) the cardinality of each partition; (2) what
objects must be, or must not be, in the same partition; and (3) which subset of objects
must be in which subset of partitions. The two latter types of constraints can be expressed by formulating restrictions
on object pairs in $\mathcal{U}$, while the first type of constraint can be specified as a cardinality vector of size $R$. Finally, $p$ is the probability of a
\emph{convergent request} \cite{gale1990}, i.e., the probability that a
request (i.e., an observation) encompasses two objects from the same underlying partition. A request
where the objects originate from different underlying partitions is called a
\emph{divergent request} \cite{gale1990}, which occurs with probability $1-p$.

Under the above model, an observation can be simulated by sampling from a Bernoulli distribution.
With probability $p$, select a pair of objects randomly from $U$:
$(O_i,O_j) \in \mathcal{U}, i\not=j$ (a \emph{convergent} request). And with
probability $1 - p$, randomly select a pair of objects not in $U$: $(O_k,O_m) \not\in \mathcal{U},
k\not=m$ (a \emph{divergent} request). This definition is equivalent to the
definition given by Oommen et al. \cite{Oommen1988}.

Previously, only heuristic sub-optimal solution strategies have been proposed for SO-EPP, and no solution exists for CSO-EPP.
In this paper, for both of these problems, we propose the first {\it optimal}  solution strategy. The solution strategy is based on a novel Bayesian network representation 
of CSO-EPP problems. To enable  swifter computations with BN-EPP, we additionally
introduce Walk-BN-EPP, an approximate reasoning approach that takes advantage of the unique structure of BN-EPP. The paper contribution can be summarized as follows:
\begin{enumerate}
	\item{We propose a novel Bayesian network model of the CSO-EPP problem
            (BN-EPP) that fully captures the nuances of CSO-EPP.}
	\item{We provide a BN-EPP based algorithm for on-line object partitioning
            that outperforms the existing \emph{state-of-the-art} SO-EPP
            solution schemes.}
	\item{The BN-EPP scheme is highly flexible in the sense that we can encode arbitrary partitioning constraints.}
	\item{Our scheme is parameter-free, which means that optimal performance
	is obtained without any fine tuning of parameters.}
    \item{In addition to predicting the optimal partitioning  of objects, BN-EPP also estimates the noise parameter $p$ on-line.}
	\item{We demonstrate that Walk-BN-EPP exhibits state-of-the-art
	performance on a large-scale real-world warehouse order picking problem.}
\end{enumerate}

The paper is organized as follows. In Sect. 2 we present related
work. We then provide a brief overview of Bayesian networks in Sect.
3, before we proceed with providing the details of our BN-EPP scheme in
Sect. 4. Then, in Sect. 5, we present our empirical results, demonstrating
the superiority of BN-EPP when compared to existing state-of-the-art schemes.
We conclude in Sect. 6 and provide pointers for further work.

%
%
\section{Related Work}
\label{sec:state-of-art}
The OPP is already a thoroughly studied problem \cite{xu2005,berkhin2006}. 
Yet, research on its fascinating variant, SO-EPP \cite{hammer1976index, adaptive1981clustering, Oommen1988, gale1990}, is surprisingly sparse 
despite its many real-world applications, which includes software clustering \cite{mamaghani2009clustering} and keyboard layout optimization \cite{oommen1991adaptive}. To cast further light on the unique properties of SO-EPP, we will here relate it to two similar problems, namely, the \emph{Poset Ordering Problem} (POP) and the \emph{Graph Partitioning Problem} (GPP).

{\bf The Poset Ordering Problem (POP).} A \emph{poset} is defined as a set of elements with a transitive partial order, where some elements may be
incomparable \cite{daskalakis2011}. A binary relation that is reflexive, antisymmetric, and transitive defines this ordering, referred to as a
\emph{less-than-or-equal }relation ($\le$). The standard \emph{less-than-or-equal } relation for integers
forms for instance a partial ordering on the set of integers.
In the poset ordering problem, the goal is to establish the partial ordering of a poset by comparing pairs of elements, typically using the\emph{ less-than-or-equal }relation as few times as possible. Accordingly, both in SO-EPP and POP, one must learn from paired elements to uncover an underlying more complex structure. That is, in POP, the \emph{less-than-or-equal} relation is applied iteratively on pairs of elements, while in SO-EPP a  \emph{in-the-same-partition} relation is used instead. Whereas the \emph{less-than-or-equal} relation found in POP is both reflexive and transitive, it is not symmetric, i.e., $A \leq B$ does not imply $B \leq A$. The \emph{in-the-same-partition} relation, however, is symmetric. This means that the solution of SO-EPP is not a partial ordering, but a set of \emph{equivalence classes}, leading to unique solution schemes.

{\bf The Graph Partitioning Problem (GPP).} The GPP is in its most general
form an NP-complete problem \cite{andreev2006}: Let $G = (V,E)$ be a graph
with a set of vertices $V$ and a set of weighted edges $E$. In graph
equipartitioning, the goal is to partition $V$ into $k$ subsets $V_1, V_2, \ldots,
V_k$ of equal cardinality. In all brevity, the solution to a GPP instance is the partitioning
that minimizes the sum of those edge weights that cross different vertex sets, $(V_i, V_j), i \neq j$ \cite{burkard2013quadratic}.  The SO-EPP can thus be cast as a
GPP if the frequencies of object co-occurrence are known for all object pairs.
Then we could form a complete graph, $G = (V,E)$, where each vertex in $V$
represents an object. Further, the weight of an edge between a pair of objects is simply
the frequency with which we observe that particular pair. The resulting GPP can then be
solved by any GPP solver \cite{galinier2011efficient,
kim2011genetic}. Since the goal is to equipartition the
graph, more specialized algorithms can also be used \cite{gupta2010}. However, in SO-EPP, the set $E$ is empty initially (due to a lack of information). Yet, after observing a sufficiently large number of object pairs, the weighted edge set, $E$, would settle down \emph{close} to the true underlying
object co-occurrence frequencies. Unfortunately, that would require an exponentially growing number of observations as the size of $V$ increases, making the GPP solvers impractical for solving SO-EPP in general.

{\bf State-of-the-art solution schemes for SO-EPP.} We now turn our attention to algorithms that are specifically designed to solve SO-EPP.
The \emph{state-of-art} solution scheme for SO-EPP is the Object Migration Automaton (OMA), introduced by Oommen et al.
\cite{Oommen1988} and later improved by Gale et al. \cite{gale1990}. The OMA is a statistics free scheme, meaning that it does
not try to estimate object co-occurrence frequencies. Instead, each object navigates a finite state machine according to a few simple fixed rules, allowing the objects to \emph{migrate}
between the different partitions, gradually converging to a solution.
While strong empirical evidence suggests that OMA asymptotically
converges to the optimal partitioning, formal convergence proofs have not yet been found, except for the trivial case of four objects and two partitions \cite{Oommen1988}.
This heuristic rule based approach, although efficient, is not optimal, which lead us to design the BN-EPP algorithm presented in this paper.
BN-EPP is a probabilistic parameter-free algorithm that, as we shall see, is not only more flexible in terms of the requirements
placed on the solution, but also able to infer the level of noise present in the environment.

\section{An Optimal Bayesian Network Based Solution Scheme for the Constrained Stochastic On-line Equi-Partitioning Problem}

In this section, we present our novel BN-EPP scheme --- a generative modeling
approach for solving CSO-EPP based on Bayesian networks (BNs). By
taking advantage of the ability of BNs to construct interpretable models
that encode probability distributions over complex domains
\cite{koller2009probabilistic}, we capture the unique characteristics of CSO-EPP. We further propose an efficient reasoning
algorithm for BN-EPP that allows uncertainty to be represented and managed
explicitly.

\begin{figure}[h!]
		\centering
		\includegraphics[width=50mm]{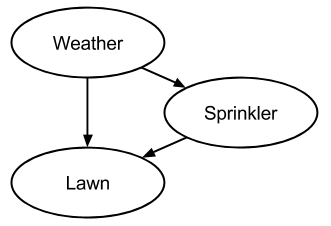}
		\captionof{figure}{Simple BN.}
 		\label{fig:bn-example}	 		
\end{figure}

A BN consists of a directed acyclic graph (DAG) representing the conditional
dependencies between a set of random variables. When modeling causal
relationships, an edge between the nodes A and B signifies that A
\emph{"causes"} B. Consider the BN shown in Figure \ref{fig:bn-example}. In this simple
BN, we have three discrete random variables: {\it Weather}, {\it Sprinkler}
and {\it Lawn}. Let us assume that the weather can have one of three different
states: {\it Sunny}, {\it Cloudy}, or {\it Rainy}. Further, the lawn is either
{\it Wet} or {\it Dry}, and the sprinkler can be {\it On} or {\it Off}. Adding
directed edges, we can encode knowledge about cause and effect, such as the
fact that rainy weather causes the lawn to be wet. Similarly, a long period of
sunny weather triggers a need for turning the sprinkler on, hence weather
indirectly causes the lawn to be wet though the sprinkler system.

The above qualitative description of cause and effect is further enriched with a
quantitative description. The quantitative description takes the form of a probability distribution assigned to each node, conditioned on
the state of the parents of the respective node. The purpose of the
conditional probability distributions is to quantitatively describe the
cause and effect relationships captured by the DAG. We assign these
probabilities though Conditional Probability Tables (CPTs), one for each node
in the graph. Note that a node without parents is assigned an unconditional probability distribution. A CPT for the sprinkler can be seen in Table \ref{tab:sprinkler-cpt}, where the effect weather has on the state of the sprinkler is captured. The CPT here tells us, e.g., that the sprinkler turns on with probability $0.8$ in sunny weather.

From the BN CPTs, we can conduct diagnostic, predictive and inter-causal reasoning, simply by asking questions about
the state of the random variables. One could for instance ask: "if the lawn is wet, what are the
chances that it was caused by rain or by the sprinkler?" or "if the
sprinkler is on, does that indicate that there is sun outside?".

\begin{table}
		\centering
		\begin{tabular}{|c|c|c|c|}
			\hline
			Weather -- State              & Sunny & Cloudy & Rainy \\ \hline
			$P(\mbox{Sprinkler} = \mbox{on} | \mbox{Weather})$ & 0.8   & 0.15   & 0.05  \\ \hline
			$P(\mbox{Sprinkler} = \mbox{off} | \mbox{Weather})$ & 0.2   & 0.85   & 0.95  \\ \hline
		\end{tabular}
		\captionof{table}{CPT of Sprinkler}
		\label{tab:sprinkler-cpt}
\end{table}



\begin{figure}[t!]
	\centering
	\includegraphics[width=70mm]{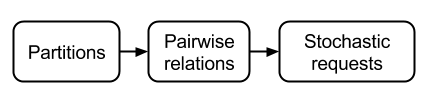}
	\captionof{figure}{Overview of BN-EPP.}
 	\label{fig:bn-epp-overview}
\end{figure}
The generative model that we propose, BN-EPP, can be described in terms of three interacting BN fragments, as shown in Figure \ref{fig:bn-epp-overview}. Firstly, a dedicated BN fragment, referred to as "Partitions" in the figure, captures the actual placement of objects into partitions. This includes any constraints on the partitioning, such as equi-partitioning. Since the objects arrive in pairs, we need to further generate an intermediate BN fragment --- the "Pairwise relations" fragment --- that explicitly extracts all pairwise object relations from the "Partitions" fragment. Finally, an observation model is derived from "Pairwise relations", capturing generation of convergent and divergent request. This latter fragment, "Stochastic requests", is based on the noise parameter $p$ and the "Pairwise relations" fragment.

Based on the BN-EPP, our on-line solution strategy for CSO-EPP can be summarized as follows. In operation, arriving object pairs (requests) are entered into the "Stochastic requests" part of the BN-EPP as observations (evidence). From these observations, pairwise relations are inferred in the intermediate fragment, which, finally, leads to a probability distribution over allowed partitions of objects in the "Partitions" fragment. Every object pair observed provides new information, and gradually, with successive observations, the probability distribution over object partitions converges to a single partitioning that solves the underlying CSO-EPP.

\begin{algorithm}[]
    \KwData{Objects $\mathcal{O} = \{O_1, O_2,\ldots, O_w \}$; Partitions $\mathcal{P} = \{P_1, P_2, \ldots, P_R\}$; and Noise resolution $N$}

    \KwResult{A BN-EPP model $\beta$: Noise $p_{\beta}$; Partitions $\mathcal{O}_{\beta}$; Pairwise relations $\mathcal{A}_{\beta}$; Stochastic requests $\mathcal{X}_{\beta}$}

    \tcc{Create partitions fragment $\mathcal{O}_{\beta}$.}
    $\mathcal{O}_{\beta} := \emptyset$\\
    \For{$i := 1$ to $W$}{
	    \tcc{$O_{\beta_i}$ assigned to a partition in $\mathcal{P}$, given preceding assignments $\mathcal{O}_{\beta}.$}
    	$O_{\beta_i}$ := Node(States=$[P_1, P_2, \ldots, P_R]$, Parents=$\mathcal{O}_{\beta}$, Distr=$F_{O_{\beta i}}$)\\
    	$\mathcal{O}_{\beta} := \mathcal{O}_{\beta} \cup O_{\beta_i}$
    }

    \tcc{Create pairwise relations fragment.}
    $\mathcal{A}_{\beta} := \emptyset$\\
    \For{$i, j \in [W \times W]$ s.t. $i < j$}{
        \tcc{$A_{\beta_{ij}}$ is true if and only if $O_{\beta_i}$ and $O_{\beta_j}$ are in the same partition.}    
    	$A_{\beta_{ij}}$ := Node(States=[True, False], Parents=$\{O_{\beta_i}, O_{\beta_j}\}$, Distr=$F_{A_{\beta}}$)\\ 
    	$\mathcal{A}_{\beta} := \mathcal{A}_{\beta} \cup A_{\beta_{ij}}$\\
    }
    
    \tcc{Create stochastic requests fragment.}

    $p_\beta$ := Node(States=$[\frac{0}{N}, \frac{1}{N}, \ldots,\frac{N}{N}]$, Parents=$\emptyset$, Distr=$\mathrm{F}_{p_\beta}$) \tcp{Noise probability.}

    $\mathcal{X}_{\beta} := \emptyset$\\

    \For{$i, j \in [W \times W]$ s.t. $i < j$}{
    	$X_{\beta_{ij}}$ := Node(States=$[\mathbb{N}_0]$, Parents=$\{A_{\beta_{ij}}, p_\beta\}$, Distr=$F_{X_{\beta}})$ \tcp{Pair observation count.}
    	$\mathcal{X}_{\beta} := \mathcal{X}_{\beta} \cup X_{\beta_{ij}}$\\
    }

	\caption{Constructing BN-EPP}
	\label{alg:build-bn-epp}
\end{algorithm}

The detailed construction of BN-EPP is outlined in Algorithm \ref{alg:build-bn-epp}. The BN-EPP needs to:
\begin{itemize}
\item[]{\bf Requirement 1} Handle constraints, such as only considering partitions of equal cardinality.
\item[]{\bf Requirement 2} Infer whether two objects belong to the same partition.
\item[]{\bf Requirement 3} Correctly handle both converging and diverging requests.
\item[]{\bf Requirement 4} Encode the actual object partitioning. 
\end{itemize}
We will now explain how the BN-EPP algorithm fulfills the above requirement. First of all, recall that $\mathcal{O} = \{O_1, O_2,$ $\ldots, O_w \}$ is  a set of $W$ objects. These are to be partitioned into $R$ different
partitions $\mathcal{P} = \{P_1, P_2, \ldots, P_R\}$. The aim is to find
some unknown underlying partitioning of the objects based on noisy observations of object pairs (convergent and divergent requests).

\noindent {\bf (Requirement 1) Only consider partitions that fulfill governing constraints (Lines 1-5)}\\
\noindent The first part of the algorithm builds the "Partitions fragment" from Figure \ref{fig:bn-epp-overview}. Briefly stated, we represent each EPP object, $O_i \in \mathcal{O}$, using a
corresponding BN node, $O_{\beta i}$. Each BN node, $O_{\beta i} \in \mathcal{O}_{\beta}$, has one state per
partition, $P_i \in \{P_1, \ldots, P_R\}$, representing the partition assigned to  $O_i$. For
instance, if we have two partitions then there will be two states per object,
one for partition $P_1$ and one for partition $P_2$.

\begin{figure}[htbp!]
	\centering
   		\includegraphics[width=40mm]{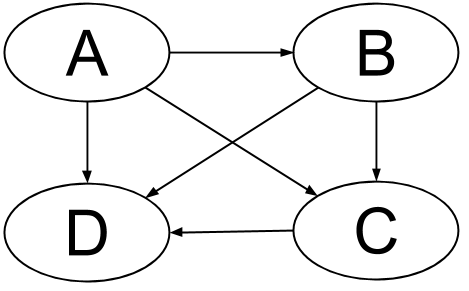}
			
		\captionof{figure}{Object dependencies for 4 objects with 2 partitions.}
		\label{fig:22-graph-objects-only}
\end{figure}

We now model the governing constraints, including equal cardinality of
partitions, by means of the BN DAG. Because of the reciprocal relationships
among objects (objects are either in the same partition or not), we can
order the BN object nodes arbitrarily. Without loss of generality,
assume that A is the first BN node in the ordering. This means that A can be
freely placed in any partition (the placement does not depend on the placement
of any other object, because none of the other objects have been placed yet).
Then the next object in the ordering, object B, only needs to take into
account object A's choice of partition. Likewise object C, the third object,
is only restricted by the previous objects' choice of partition (the choices
of object A and B). Continuing in this manner, we can always represent the
partition of the next object as solely being dependent on the already
partitioned objects. It is for this purpose we maintain the gradually increasing object set $\mathcal{O}_{\beta}$, containing all the already partitioned predecessor objects.
This organization of objects is thus leading to a BN DAG structure, as exemplified in Figure
\ref{fig:22-graph-objects-only}, capturing two partitions and four objects.

The corresponding CPTs for the EPP objects ($F_{O_{\beta i}}$ in the algorithm) are generated as a function of the
constraints set by CSO-EPP (the constraints governing the partitioning, e.g., equi-partitioning).
As an example, the CPT of object C (the third object) can be seen in Table
\ref{tab:object-c-cpt}. From the table we observe for instance that $P(C=P_1 |
A=P_1, B=P_1) = 0.0$, that is, if object A and B is in
$P_1$ then the probability of C being in $P_1$ is zero. On the other hand, if
object A and B is located in different partitions then object C is equally
likely to be in partition $P_1$ as in partition $P_2$. Thus, by constructing
the CPT of each node (representing an object) in this manner, a solution that
fulfills all of the constraints is always ensured because a partitioning that
violates constraints is assigned a probability of zero.

\begin{table}
	\centering	
		\begin{tabular}{|c|c|c|c|c|}
			\hline
			A - State & \multicolumn{2}{|c}{$P_1$} & \multicolumn{2}{|c|}{$P_2$} \\ \hline
			B - State & $P_1$         & $P_2$         & $P_1$          & $P_2$         \\ \hline
			C in $P_1$   & 0.0        & 0.5        & 0.5         & 1.0        \\ \hline
			C in $P_2$          & 1.0        & 0.5        & 0.5         & 0.0        \\ \hline
		\end{tabular}
		\captionof{table}{CPT of object C}
		\label{tab:object-c-cpt}
\end{table}

\noindent {\bf (Requirement 2) Infer pairwise relations between the objects (Lines 6-10)}\\
Now that the "Partitions fragment" has determined the partition of each object,
it is a simple task to determine whether an object pair belongs to the same partition. In the "Pairwise relations" fragment,
we represent every pair of objects as a deterministic node with two states: \emph{True} if the pair is in the same partition,
and \emph{False} when they are not (distribution $F_{A_{\beta}}$ in the algorithm).

\begin{figure}[htbp]
	\centering
   		\includegraphics[height=50mm]{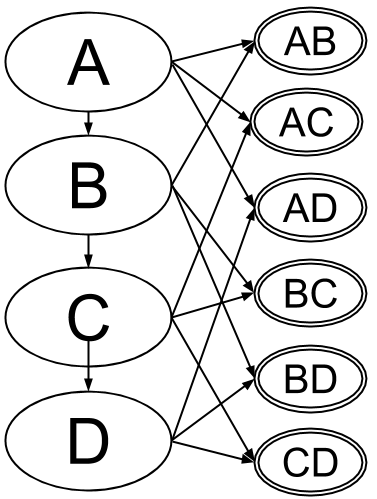}
		\captionof{figure}{Pairwise object relations for 4 objects with 2 partitions.}
		\label{fig:22-graph-pairs}
\end{figure}

Figure \ref{fig:22-graph-pairs} provides an example of a "Pairwise relations" fragment, obtained following the above procedure for four objects and two partitions. The corresponding CPT for
the pair node for object A and object C (node AC) can be found in Table \ref{tab:22-truth-table}. From
the truth table it is evident that if object A and C belong to the same partition, the state of node AC state is \emph{True},
and \emph{False} otherwise.

\begin{table}[htbp]
\centering	
		\begin{tabular}{|c|c|c|c|c|}
		\hline
		A - State  & \multicolumn{2}{|c}{$P_1$} & \multicolumn{2}{|c|}{$P_2$} \\ \hline
		C - State  & $P_1$         & $P_2$         & $P_1$          & $P_2$         \\ \hline
		AC - State & True       & False      & False       & True       \\ \hline
		\end{tabular}		

 		\captionof{table}{Truth-table for node AC}
 		\label{tab:22-truth-table}
\end{table}

The "Pairwise relations" fragment gives BN-EPP the capability to infer object
relations from pairwise observations, such as in the following scenario:
Given the above example, assume that we know that (1) Object A is known to be in partition $P_1$, and (2) Object B and
object D should be in the same  \emph{unknown} partition, i.e., the BD-node is
set to \emph{True}. BN-EPP will then correctly infer the only possible partitioning, namely the two partitions $P_1:\{A, C\}$
and $P_2:\{B, D\}$. While similar result could have been obtained though the usage of a
propositional logic solver, as we shall see, the stochastic nature of CSO-EPP
rules out such a solution.

\noindent {\bf (Requirement 3) Stochastic requests (Lines 11-15)}\\
The BN model obtained through the "Partitions"- and "Pairwise relations" fragments allows us to infer the correct object partitions, given that we know the state of a sufficient number of the pairwise relation nodes. However, the CSO-EPP involves both convergent
and divergent requests. Consequently, we need a mechanism for handling noisy information.

Firstly, we introduce a BN node $p_\beta$ representing $p$ --- the convergent request probability. The state space of $p$ is
a discretization of potential values for $p$, each with an equal prior probability. Attached to this $p_\beta$ node is a
series of \emph{observation} nodes $X_{\beta_{ij}} \in \mathcal{X}_\beta$, each dependent on the state of the $p_\beta$ node, and whether or not its
corresponding pair node $A_{\beta ij}$ is \emph{True} or \emph{False}.
The CPT for each observation node ($F_{X_{\beta}}$ in the algorithm) is a function of the number times $n \in \mathbb{N}_0$ that particular pair has been 
observed, as well as the states of the $p_\beta$ node, as shown in Table \ref{table:cpt-observation-node}. As seen, $F_{X_{\beta}}$ is distributed according to a Bernoulli distribution, $B(n,p)$. 

\begin{table}[]
\centering
\caption{The CPT of an observation node conditioned on the state of the parent pair $A_{\beta_{ij}}$ and the $p_\beta$ node.}
\label{table:cpt-observation-node}
\begin{tabular}{|l|c|}
\hline
$P(X_{\beta ij} = n | A_{\beta_{ij}}  = \mathrm{True}, p_\beta = p)$  &  $B\left(n, p \cdot \frac{1}{P} \cdot {{\frac{W}{P}}\choose{2}}  \frac{1}{\frac{W}{P}} \cdot \frac{1}{\frac{W}{P} - 1}\right)$   \\ \hline
$P(X_{\beta ij} = n | A_{\beta_{ij}}  = \mathrm{False}, p_\beta = p)$ & $B\left(n, (1-p) \cdot {{P}\choose{2}} \frac{1}{P} \cdot \frac{1}{P-1} \cdot \frac{1}{\frac{W}{P}} \cdot \frac{1}{\frac{W}{P}}\right)$ \\ \hline
\end{tabular}
\end{table}

\begin{figure}[htbp]
{
  \centering
  \includegraphics[width=80mm]{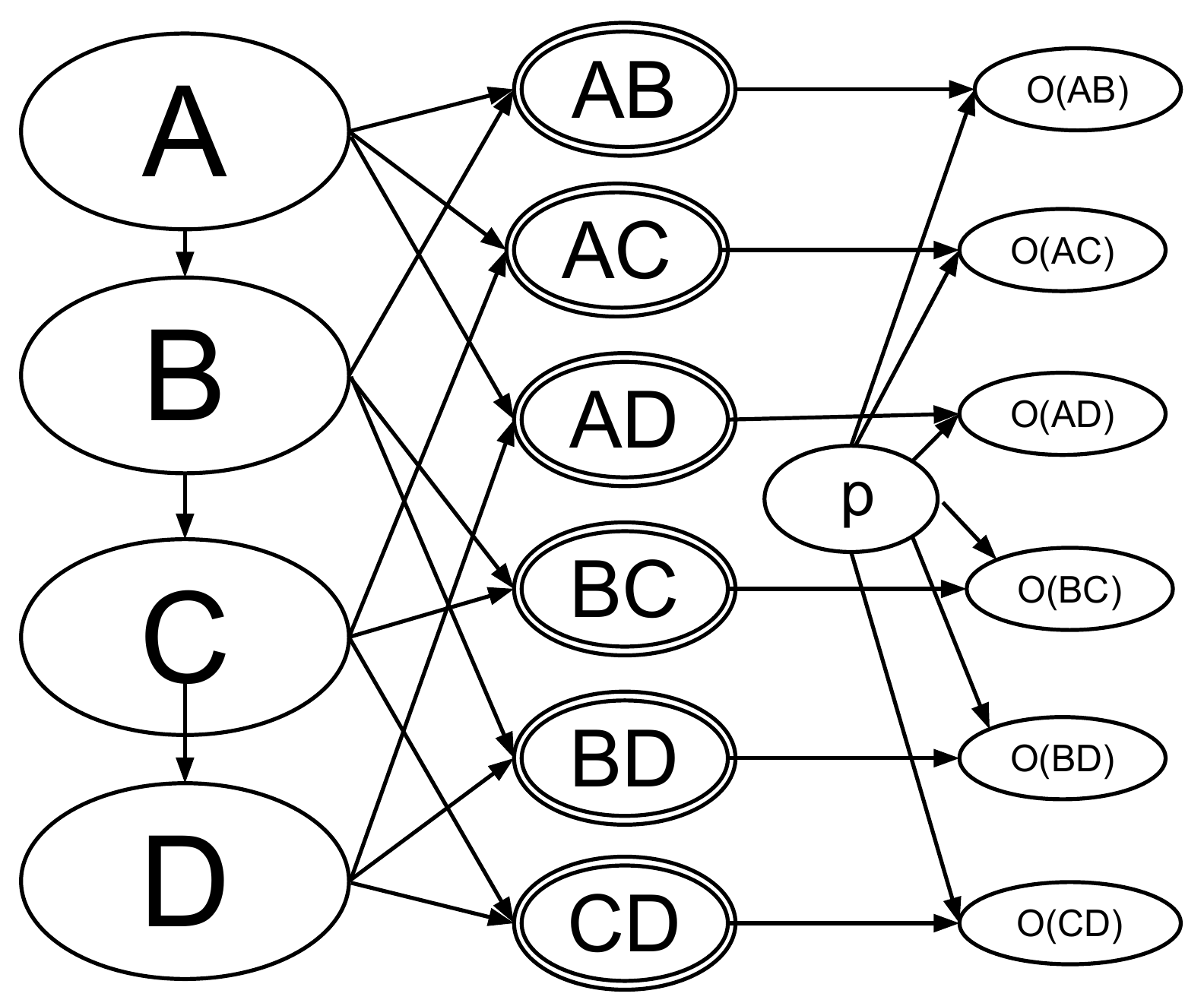}
  \captionof{figure}{BN for solving an EPP with 4 objects, 2 partitions, and Binomially distributed observation nodes.}

  \label{fig:22-graph}
}
\end{figure}

\noindent {\bf (Requirement 4) Decode the object partitioning from the BN representation}\\
While the BN correctly models the CSO-EPP, it does not directly present us with a solution in the form of a partition
for each object. However, we obtain the partitioning indirectly by finding the Maximum a Posteriori (MAP\footnote{Also known as Most Probable Explanation (MPE).}) configuration of the BN-EPP. In all brevity,
a MAP query identifies the most probable solution given the observations \cite{yuan2004annealed, koller2009probabilistic}.
\[
	\mbox{solution(BN)} = \mbox{MAP}(\mathcal{O}_\beta \cup  \mathcal{A}_\beta \cup p_\beta | \mathcal{X}_\beta) = \argmax_{\mathcal{O}_\beta \cup  \mathcal{A}_\beta \cup p_\beta} P(\mathcal{O}_\beta \cup  \mathcal{A}_\beta \cup p_\beta | \mathcal{X}_\beta)
\]
For an example of the outcome of the final step, see Figure \ref{fig:22-graph}. Note that the observation nodes for an object pair $XY$ in the figure is denoted by $O(XY)$. The complete BN-EPP for four objects and two partitions is shown, ready for MAP inference. As can be seen, the resulting BN-EPP has a complex structure. In the next section we take advantage of this structure to propose an efficient and novel inference algorithm for large scale CSO-EPP problems.

Note that the BN-EPP solution strategy is related to the Thompson Sampling (TS) principle that was introduced
by Thompson in 1933 \cite{thompson1933}, and forms the basis for several of
the leading solution schemes for so-called Multi-Armed Bandit (MAB) Problem. 
The classical MAB problem is a sequential resource allocation problem. At each
time step, one pulls one out of multiple available bandit arms. Each arm
pulled provides a reward with a certain probability, and the objective is to
maximize the total number of rewards obtained through the sequence of arms
pulled \cite{bubeck2012regret,granmo2010tsla}. In the Learning Automata (LA) literature this scheme is
referred to as Bayesian Learning Automata (BLA) \cite{granmo2010tsla}.

In TS, to quickly shift from exploring reward probabilities to reward
maximization, one recursively estimates the reward probability of each arm
using a Bayesian filter. To determine which arm to play, one obtains a reward
probability sample from each arm, and the arm that provides the highest value
is pulled. The selected arm triggers a reward, which in turn is used to
perform a Bayesian update of the arm's reward probability estimate. As a result,
TS selects arms with a frequency proportional to the posterior probability
that the arm is optimal.\cite{granmo2010tsla}

TS has turned out to be among the top performers for traditional MAB problems
\cite{granmo2010tsla,chapelle2011empirical}, supported by theoretical regret
bounds \cite{agrawal2011analysis,agrawal2013further}.
It has also been been successfully applied to contextual MAB problems
\cite{agrawal2013thompson}, Gaussian Process optimization
\cite{glimsdal2013gaussian}, Distributed Quality of Service Control in Wireless
Networks \cite{granmo2013accelerated}, Cognitive Radio Optimization
\cite{jiao2015optimizing}, as well as a foundation for solving the Maximum a
Posteriori Estimation problem \cite{tolpin2015maximum}.

%
%

\section{Walk-BN-EPP}

The MAP problem is NP-complete \cite{koller2009probabilistic}, and thus cannot
be solved efficiently for large networks in general. Accordingly, to
allow solutions to be found for large CSO-EPPs, we will in this section
introduce a novel inference scheme, Walk-BN-EPP.  Walk-BN-EPP is designed to
take advantage of the particular characteristics of the BN-EPP DAG structure and is based on WalkSAT \cite{selman1994noise}, a well-known and effective
solver for the NP-complete Boolean satisfiability (SAT) problem.

Note that our decision to design a dedicated algorithm for BN-EPP
does not mean that existing general MAP solvers, such as Variable Elimination,
Belief Propagation and the various evolutionary algorithms
\cite{larranaga2013review}, cannot be used. On the contrary, they work quite well on small- and medium sized CSO-EPPs.
However, since they do not take
advantage of the BN-EPP's unique structure, they scale poorly. Thus, by introducing Walk-BN-EPP we expand the class of problems that can be solved with BN-EPP. 

Walk-BN-EPP is based on WalkSAT \cite{selman1994noise}, a successful
algorithm for solving the NP-complete Boolean satisfiability (SAT) problem
\cite{soh2017proposal}. In all brevity, in SAT the goal is to find a truth value
assignment for the variables of a Boolean expression that makes the
overall expression  evaluate to "True", thus \emph{satisfying} the expression.
The Boolean expression is a propositional logic formula that consists of a
conjunction of Boolean clauses. The overall strategy of WalkSAT can be
summarized as follows. One repeatedly selects one of the Boolean variables randomly,
negate its value, and then observe whether the new truth value increases the
total number of Boolean clauses satisfied. If the number of satisfied clauses
does not increase, then with high probability one reverts the negated Boolean
variable to its original state. Otherwise, the new state is kept. This simple iterative procedure is repeated
until all the clauses are satisfied. Hence, one could say that WalkSAT
performs a random walk with a drift towards "better" truth value assignments,
that is, assignments with an increasing number of clauses satisfied.

Walk-BN-EPP is inspired by Walk-SAT in the sense that we divide
Walk-BN-EPP into two steps: (1) Generate an initial configuration that 
partitions the objects by sampling from BN-EPP using forward sampling. (2) Improve the initial partitioning
by applying a random walk with a drift towards more probable partitionings,
that is, BN variable state configurations with higher MAP. The two steps are laid out in Algorithm \ref{alg:walk-bn-epp}, and we here
explain them in more detail.

{\bf Initialization Step.} In order to perform a Walk-SAT inspired random walk, we need an initial state configuration for the variables in BN-EPP. This initial configuration should ideally be as close as possible to the solution we seek, to reduce the length of the random walk. To achieve this, we sample an initial configuration from a rough estimate of the posterior probability distribution, one object $O_{\beta_i}$ at a time, starting with $O_{\beta_1}$. That is, the state assigned to $O_{\beta_i}$ is sampled from $P(O_{\beta_i} | \mathcal{X}_{\beta}, \{O_{\beta_k}\}^{i-1}_{k=1})$ using the traditional Likelihood-Weighted (LW) sampling algorithm \cite{koller2009probabilistic}. Since the "Pairwise relations" fragment follows deterministically from the "Partitions" fragment, the states of the nodes $\mathcal{A}_\beta$ are then also given. When all of the object nodes, $O_{\beta_i} \in \mathcal{O}_\beta$, have been assigned a state in this manner, we use this configuration as an initial solution candidate for the random walk. The details of the initialization step are covered by lines 1-5 in Algorithm \ref{alg:walk-bn-epp}. 

Note that constraints forces the posterior probability of any violating assignment to zero, with the remaining probabilities renormalized.  As an example, assume that we have 16 objects and 4 partitions. We have already placed 4 objects
into partition number 3. To place the 5th object, use LW sampling and obtain $P(O_{\beta_5}) = \{0.1, 0.7, 0.2, 0.0\}$ from the BN. Note that the fourth probability becomes zero due to the previous assignment of objects to the corresponding partition, reflecting a full partition. To place the 5th object we then sample a partition from $P(O_{\beta_5})$. That is, we select partition 1 w.p. $0.1$,
partition 2 w.p. $0.7$ and partition 3 w.p. $0.2$. In this example, let us assume that we sampled partition 1. The 5th object is thus assigned to this partition. We repeat this process for each object, taking into account the choices
of all previously assigned objects, until all objects have been assigned to a partition.

{\bf The Walk-SAT Based Search.} In the second step of our algorithm (lines 6-22), we seek to iteratively improve the initial configuration from the initialization step. We do this by performing a Walk-SAT inspired random walk over the state space of candidate partitions. The random walk consists of iteratively swapping the partition of randomly selected pairs of objects, $(O_{\beta_i}, O_{\beta_j}) \in \mathcal{O}_\beta \times \mathcal{O}_\beta $, with the intent of gradually moving towards more probable object partitions, and ultimately, the most probable partitioning (i.e., the solution to the MAP problem). Let the set $\mathcal{O}^t_\beta = \{O_{\beta_1} = o_1, O_{\beta_2} = o_2, \ldots, O_{\beta_i} = o_i, \ldots, O_{\beta_j} = o_j, \ldots, O_{\beta_n} = o_n \}$ be the current configuration of the network before two randomly selected objects, $O_{\beta_i}$ and $O_{\beta_j}$, swap partitions. Further, let $\mathcal{O}^{t+1}_\beta = \{O_{\beta_1} = o_1, O_{\beta_2} = o_2, \ldots, O_{\beta_i} = o_j, \ldots, O_{\beta_j} = o_i, \ldots, O_{\beta_n} = o_n \}$ be the configuration produced by the swap. Finally, let the log probability, $C^q$, of a configuration $q$ be defined as follows:
\[
    C^q = \log P(\mathcal{O}^q_\beta) = \sum_{1 \le k \le N} \log P(O_{\beta_k} = o_k | \mathit{parents}(O_{\beta_k}))
    \label{eq:log-prob}
\]
with $\mathit{parents}(O_{\beta_k})$ being the parents of the node $O_{\beta_k}$ in BN-EPP.

To systematically refine the current configuration, we always switch from configuration $\mathcal{O}^t_\beta$ to 
configuration $\mathcal{O}^{t+1}_\beta$ if the log probability $C^t$ is
greater than $C^{t+1}$ (we accept the new configuration).
If, on the other hand, the log probability decreases, we instead reject the new configuration, $\mathcal{O}^{t+1}_\beta$, with probability $1-\epsilon$. Otherwise, we accept the new configuration. Note that in the algorithm, $U(0,1)$ refers to a uniform distribution over the interval $[0,1]$..

As an example assume that $C^4 = -15.3$, we then pick one objects from two different partitions, 
say object number 4 and 10 and swap their location. Calculating $C^5 = -14.9$ we observe that
$C^5$ is greater than $C^4$, thus we accept the new state $\mathcal{O}^{5}_\beta$.
For the next step, we select object 1 and 2 and swap their locations. However, calculating $C_6 = -20.2$ we
see that the previous state, $\mathcal{O}^5_\beta$, has a larger log probability than the new configuration. Therefore we revert to the original configuration with probability $1-\epsilon$, else, w.p. $\epsilon$ we
keep the new, though inferior configuration. This process is then repeated for a predefined number of steps $T$ and the best
observed configuration is presented as the solution.

\begin{algorithm}[]
    \KwData{Bayesian network BN-EPP, $\epsilon$ - probability of
            accepting an inferior state, and $T$ - the number of steps to execute.}
    \KwResult{MAP configuration}
    
    \For{$i := 1$ to $W$}{
    	Estimate $\pi_i = P(O_{\beta_i} | \mathcal{X}_\beta, \{O_{\beta_k}\}^{i-1}_{k=1})$ using LW.\\
    	Draw a single sample from $\pi_i$: $s \sim \pi_i$.\\ 
    	Set the state of object $i$: $O_{\beta_i} = s$\\
    }
    $\mathcal{O}^0_\beta = \{O_{\beta_1} = o_1, O_{\beta_2} = o_2, \ldots, O_{\beta_n} = o_n \}$\\
    $C_0 = \textrm{CalculateLogProbability}(\mathcal{O}^0_\beta)$\\
    $\mathcal{O}^{max}_\beta := \mathcal{O}^0_\beta$\\
    $C^{max} := C^0$\\
    
    \For{$t := 1$ to $T$}{
        
        $O_{\beta_i}, O_{\beta_j}$ = \textrm{PickTwoRandomObjects()}\\
        $\mathcal{O}^t_\beta$ := \textrm{SwapPartitionsOfObjects($O_{\beta_i}, O_{\beta_j}, \mathcal{O}{t-1}_\beta$)}\\
        $C^t$ := \textrm{CalculateLogProbability($\mathcal{O}^t_\beta$)}\\
        \vline

        \If{$C^{max} < C^t$}{
            $\mathcal{O}^{max}_\beta := \mathcal{O}^t_\beta$\\
            $C^{max} := C^t$\\
        }
        
        \If{$C^t < C^{t-1} \mbox{ and } U_{(0,1)} < 1-\epsilon$}{ 
            $\mathcal{O}^t_\beta$ := $\mathcal{O}^{t-1}_\beta$\\
            $C^{t} := C^{t-1}$
        }
    }

    \Return{$\mathcal{O}^{max}_\beta$}

  \caption{Walk-BN-EPP}
  \label{alg:walk-bn-epp}
\end{algorithm}

%
%
\section{Experimental Results on Walk-BN-EPP}
\label{sec:results}
To evaluate the on-line performance of BN-EPP and Walk-BN-EPP, we will here study convergence speed and accuracy empirically.
The main question is how many observations, or queries, are required to obtain a correct partitioning of the objects, for various stochastic environments. Since the response to queries is stochastic, we will measure average performance over a large ensemble of independent trials. 

The different stochastic environments that we investigate are found in Table \ref{tab:p-for-env}, which also lists the log probability for generating the correct partitioning by chance, for each environment. A lower probability indicates that it will be more difficult to find the correct partitioning.  Observe that it is significantly harder to find the correct partitioning for the stochastic environment with three partitions and nine objects (abbreviated r3w9), with r4w16 being even more challenging. Note further that previous work has mostly been concerned with the r2w4 and r3w9 problems, so these problems will serve as a benchmark when we compare our approach with state-of-the-art, while r4w16 will be used to study the scalability of Walk-BN-EPP.

\begin{table}[htbp]
    \centering
        \begin{tabular}{|l|c|c|c|c|c|}
            \hline
            Problem & r2w4 & r2w6 & r3w6 & r3w9 & r4w16 \\ \hline
            log P(correct partitioning)  & -1.09    & -2.30   & -2.70   & -5.63  &  -14.78  \\ \hline
        \end{tabular}
    \captionof{table}{The probability of randomly generating a correct partitioning in the different stochastic
    environments. Here r2w4 denotes that the SO-EPP instance consists of 2 partitions with 4 objects in total. Notice how the
    EPP becomes several orders of magnitude harder to solve as we increase the number of objects and partitions.}
    \label{tab:p-for-env}
\end{table}

\subsection{Impact of Walk-BN-EPP Parameter Settings}

To evaluate the impact of the various parameters available in Walk-BN-EPP, we first solve the r4w16 problem
using a diverse range of parameter configurations. These include different number of random walk steps as well as thoroughness in sampling the initial configuration (i.e., number of samples used to estimate a maximum posterior initial configuration). Not surprisingly, as seen in Table \ref{tab:walk-bn-epp-internal-results}, increasing the number of
steps in the random walk significantly enhances the performance of Walk-BN-EPP. In addition, we observe that
increasing the number of samples used to estimate an initial configuration increases performance further. Indeed, by applying our likelihood-weighted
sampling algorithm by the modest number of 250 samples per object, we 
obtain an 1120\% increase in probability of finding the configuration that provides the maximum posterior probability.

\begin{table}[]
\centering
\caption{Walk-BN-EPP results for different configurations on the r4w16
        problem with 100 observations and p=0.75. To perform inference for the TS
        prior generator likelihood-weighted sampling was used with the number of samples
        as indicated in the table. Each data point is the average of a 1000 independent trials.
        }
\label{tab:walk-bn-epp-internal-results}
\begin{tabular}{|l|c|c|c|c|c|c|}
\hline
{\bf Walk Iterations} & {\bf 50} & {\bf 100}  & {\bf 500} & {\bf 1000} & {\bf 2000} & {\bf 4000}\\ \hline \hline
Random Prior        & 0.005 & 0.02  & 0.039 & 0.05  & 0.057 & 0.064 \\ \hline
TS with 50 samples       & 0.007 & 0.039 & 0.048 & 0.056 & 0.069 & 0.070 \\ \hline
TS with 250 samples      & 0.061 & 0.066 & 0.063 & 0.085 & 0.11  & 0.10  \\ \hline
\end{tabular}
\end{table}

\subsection{Empirical Comparison with the Object Migration Automaton (OMA)}
The Object Partitioning Automata (OMA) \cite{Oommen1988} represents state-of-the-art for solving EPP. We here compare our novel BN-EPP approach focusing on:
\begin{itemize}
\item Rate of convergence, i.e., how many requests do we need to observe before we are able to correctly
partition the objects. 
\item Probability of convergent requests (degree of noise).
\end{itemize}
For each experiment configuration, ten thousand individual trials were performed in order to minimize variance in our results.
To avoid bias, we further independently selected a random optimal partitioning of the objects for every trial, and made sure that both algorithms were exposed to an identical sequence of incoming queries.

Unlike OMA, which requires a predetermined number of states (denoted $N$ in this paper), BN-EPP is a parameter free scheme. We here report the results of OMA using the standard choice of 10 states ($N=10$) \cite{Oommen1988,gale1990} for all of the experiment configurations.
Note that the experiments were also executed with states $N=5$ and $N=20$. However, $N=10$ provided the best performance overall.

\begin{figure}[h!]
	\centering
		\includegraphics[width=70mm]{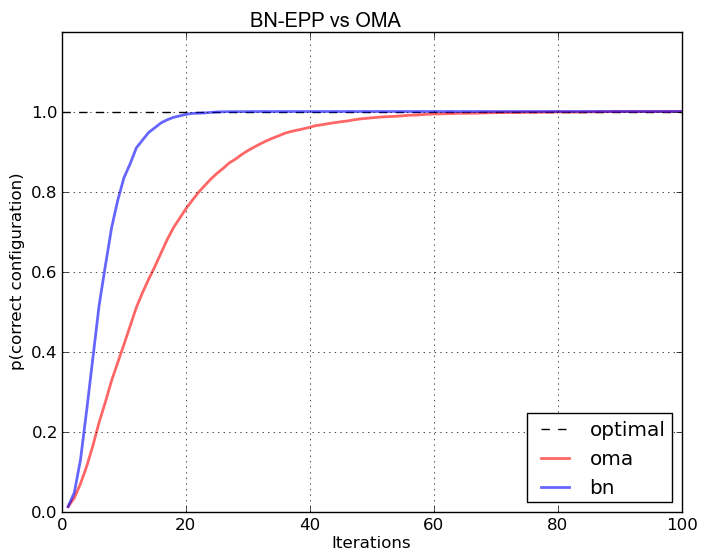}		
		\includegraphics[width=70mm]{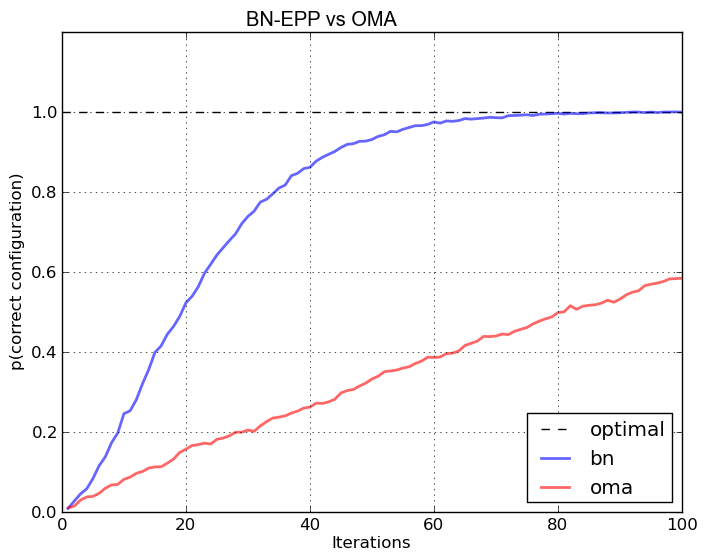}
	\captionof{figure}{BN-EPP vs OMA with $R=3$, $W=9$. The probability of convergent requests is $p=0.9$ to the left and $p=0.6$ to the right.}
	\label{fig:33}
\end{figure}

\begin{figure}[h!]
	\centering
	\includegraphics[width=70mm]{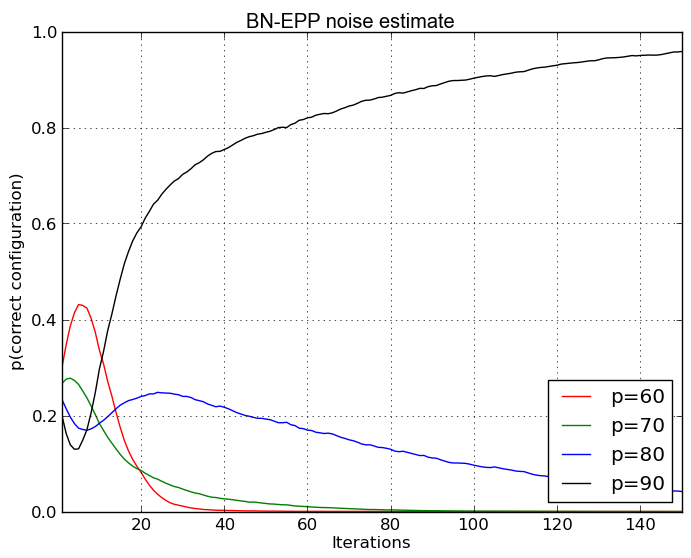}		
	\includegraphics[width=70mm]{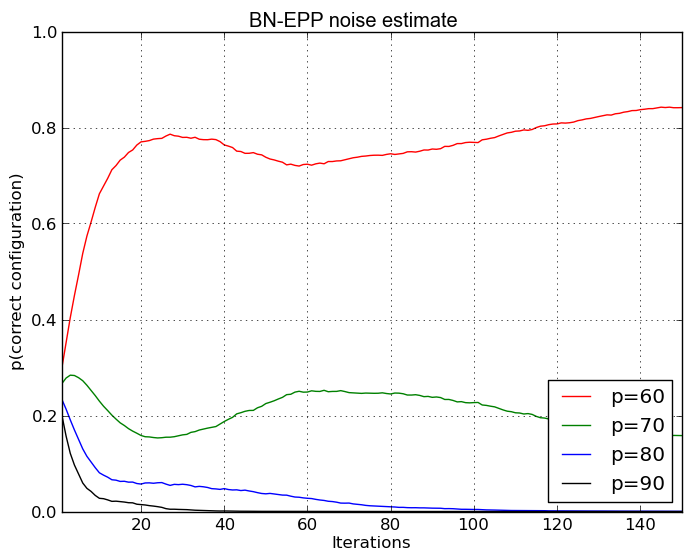}
	\captionof{figure}{Probability of the different $p$ values for $R=3$ and $W=9$. The left plot covers the scenario where the true probability of convergent requests $p$ is  $0.9$, while the right plot shows the results for convergent request probability $0.6$.}
	\label{fig:p-of-33}
\end{figure}

Representative results can be found in Figure \ref{fig:33}. In the figure, we observe that 
BN-EPP's rate of convergence greatly exceeds that of OMA. Furthermore,
the performance advantage of BN-EPP increases with level of noise.
The reasoning behind this disparity is that OMA implicitly assumes that all requests are convergent requests, and trust that \emph{"on average"} 
there will be more convergent requests than divergent requests, guiding OMA towards convergence. BN-EPP, on the other hand, directly quantifies the uncertainty associated with the requests by estimating $p$ -- the probability of a convergent request. In Figure \ref{fig:p-of-33} we have  plotted the probabilities BN-EPP assigned to the different $p$ values
from time step to time step. A major feature of BN-EPP is that it maintains a probability distribution spanning the whole object partitioning solution space, while OMA only works from a single configuration instance. We further believe that the ability of BN-EPP to track $p$ explains why BN-EPP infers the correct partitioning significantly faster than OMA.

The performance of BN-EPP and OMA in the other scenarios is summarized in Table \ref{tab:all-other}, which demonstrates a similar trend.

\begin{table}[h]
\centering
\begin{tabular}{c|c|c|}
	\cline{2-3}
	\multicolumn{1}{l|}{}      & $E[P(\mbox{correct})] \mbox{ at }t=10$    & $E[P(\mbox{correct})]\mbox{ at }t=50$      \\ \hline
	\multicolumn{1}{|c|}{r2w4} & BN-EPP=0.89, OMA=0.85 & BN-EPP=0.99, OMA=0.98 \\ \hline
	\multicolumn{1}{|c|}{r2w6} & BN-EPP=0.83, OMA=0.71 & BN-EPP=0.99, OMA=0.96 \\ \hline
	\multicolumn{1}{|c|}{r3w9} & BN-EPP=0.47, OMA=0.32 & BN-EPP=0.89, OMA=0.61 \\ \hline
\end{tabular}
\captionof{table}{BN-EPP vs OMA for different problem configurations with $p=0.6$. Here the table indicates the expected probability of success estimated from the average of a thousand independent trials.}
\label{tab:all-other}
\end{table}
 


\subsection{Empirical Results for Warehouse Optimization}
To demonstrate the applicability of BN-EPP, we evaluate our scheme using one month (30 days) of
real-world point-of-sale transaction data from a grocery outlet (collected in \cite{hahsler2006implications}).
Each transaction $T_k$ is a subset of the set of all unique articles $O$, where $|O| = 169$. In total there are $9835$ transactions. Each article is labeled by its type of product, e.g. ice-cream
instead of the actual brand. The number of articles per transaction vary wildly from orders of size $32$ down to single
article transactions. The mean number of objects per transaction is $4.4$, with a standard deviation of $3.5$.

We here assume that the objects are to be partitioned among 13 different
sections in such a manner that each customer will minimize the number of times
they travel to another section of the store when collecting the articles on
their shopping list (transaction $T_k$).

In addition, as discussed in the introduction, we introduce constraints on the placement of objects, listed in Table \ref{tab:warehouse-restrictions}.
However, as OMA does not support these
kind of restrictions, we will include results for BN-EPP, both with and without these restrictions in place.

To measure solution effectiveness, we track how many warehouse sections, $v$, a consumer must visit to collect
all the wares on his shopping list. We then assume that the experienced cost of travel doubles for each new unique section the
consumer must visit. As an example, if a customer needs to visit $3$ different sections the cost of that transaction becomes $2^3 = 8$.

We evaluate the effectiveness of BN-EPP using $5$ fold cross validation, where we select $1$ fold for training and $4$ folds
for testing. We report the mean cost of the transactions in the test set. The $5$-fold cross validation is performed $1000$ times to estimate
expected effectiveness. For Walk-BN-EPP, we used the parameter settings of likelihood-weighted sampling with 100 samples per object and 1000 iterations for the walk phase.

\begin{table}[]
\centering
\caption{Effectiveness of Walk-BN-EPP and OMA on the grocery dataset \cite{hahsler2006implications} 
 as measured in number of sections traveled (5-fold cross validation).}
\label{my-label}
\begin{tabular}{|l|c|c|c|}
\hline
        & BN-EPP (rules) & BN-EPP (no rules) & OMA (no rules)  \\ \hline
Mean    &   61.6    & 30.6           & 68.1  \\ \hline
Std.Dev &   6.1     & 4.5            & 6.5 \\ \hline
\end{tabular}
\end{table}

\section{Conclusion and Further Work}
In this paper we have presented a novel approach to the Constrained Stochastic Online Equi-Partitioning Problem (CSO-EPP), namely, the
Bayesian Network EPP model and inference scheme. We have demonstrated how the various components of BN-EPP interact and
that BN-EPP significantly outperform existing state-of-art, not only in speed
of convergence, but also in its ability to estimate the stochasticity of the
underlying environment. From a history of object arrivals, we are able to
predict which objects will appear together in future arrivals. To enable BN-
EPP to deal with larger data sets we introduced Walk-BN-EPP, a WalkSAT
inspired solver for BN-EPPs. Walk-BN-EPP was then applied to a real-world
warehouse problem and shown to significantly outperform state-of-the-art inference schemes,
even when constraining the solution space in terms of real-world constraints.

In our future work, we intend to investigate how the BN-EPP approach can be expanded to cover other classes of stochastic optimization problems such as graph partitioning and poset ordering problems, potentially outperforming generic off-line techniques such as Particle Swarm Optimization (PSO) \cite{kennedy2011particle},
Genetic Algorithm (GA) \cite{holland1992genetic} or Ant Colony Optimization (ACO) \cite{dorigo2006ant}.

\bibliographystyle{IEEEtran}
\bibliography{library}
\end{document}